\documentclass{article}

\usepackage[preprint]{neurips_2024}

\usepackage[utf8]{inputenc} 
\usepackage[T1]{fontenc}    

\usepackage{tabularx}

\usepackage{multirow}
\usepackage{makecell}
\newcommand{\ours}{SC-Pro\xspace}
\newcommand{\oursfull}{\textbf{S}pherical or \textbf{C}ircular \textbf{Pro}bing\xspace}
\newcommand{\oursone}{SC-Pro-o\xspace}

\newcommand{\dmdsd}{\(\text{DMDv2}_\text{SDv1.5}\)~}
\newcommand{\dmdsdxl}{\(\text{DMDv2}_\text{SDXL}\)~}

\usepackage[pagebackref,breaklinks,colorlinks]{hyperref} 

\usepackage{url}            
\usepackage{booktabs}       
\usepackage{amsfonts}       
\usepackage{nicefrac}       
\usepackage{microtype}      
\usepackage{xcolor}         

\usepackage{amsmath}
\usepackage{amssymb}
\usepackage{cleveref}
\usepackage{subcaption}
\usepackage{multirow}   
\usepackage{makecell}   
\usepackage{wrapfig}    
\usepackage{scalerel,stackengine}
\stackMath
\newcommand\reallywideacute[1]{%
  \ThisStyle{\savestack\tmpA{$\SavedStyle#1$}%
  \savestack{\tmpbox}{%
    \def\scriptstyleScaleFactor{0.8}
    \def\scriptscriptstyleScaleFactor{0.68}
    $\SavedStyle\stretchto{%
    \scalerel*[\wd\tmpAcontent]%
      {\kern-2.05\LMpt\mathchar"7013\kern-1.1\LMpt}%
    {\rule{0ex}{\textheight}}%
  }{2.0\LMex}$}%
  \stackengine{-7\LMpt}{\SavedStyle#1}{\tmpbox}{O}{c}{F}{T}{S}}%
}

\usepackage{listings}

\usepackage{verbatim}

\title{SC-Pro: Training-Free Framework\\for Defending Unsafe Image Synthesis Attack}

\author{
\vspace{.25em}
\textbf{Junha Park} \quad \textbf{Jaehui Hwang} \quad \textbf{Ian Ryu} \quad \textbf{Hyungkeun Park} \quad \textbf{Jiyoon Kim} \\\vspace{-.5em}
\textbf{Jong-Seok Lee} \\\\\vspace{-.5em}
Yonsei University \\\\
\texttt{\{junha.park,jaehui.hwang,ianryu,hyungkeun.park,ji-yoon.kim,jong-seok.lee\}}\\
\texttt{@yonsei.ac.kr}
}

\begin{document}

\maketitle

\begin{abstract}
With advances in diffusion models, image generation has shown significant performance improvements.
This raises concerns about the potential abuse of image generation, such as the creation of explicit or violent images, commonly referred to as Not Safe For Work (NSFW) content.
To address this, the Stable Diffusion model includes several safety checkers to censor initial text prompts and final output images generated from the model.
However, recent research has shown that these safety checkers have vulnerabilities against adversarial attacks, allowing them to generate NSFW images.
In this paper, we find that these adversarial attacks are not robust to small changes in text prompts or input latents.
Based on this, we propose \ours (\oursfull), a training-free framework that easily defends against adversarial attacks generating NSFW images.
Moreover, we develop an approach that utilizes one-step diffusion models for efficient NSFW detection (\oursone), further reducing computational resources.
We demonstrate the superiority of our method in terms of performance and applicability.
\vspace{-1em}
\end{abstract}

\noindent\resizebox{\textwidth}{!}{%
    \colorbox{yellow!30}{%
        \parbox{\dimexpr\textwidth-2\fboxsep\relax}{%
            \centering
            \textbf{\textcolor{red}{Warning!}} This document contains sensitive images.
            \\ While we censor Not-Safe-for-Work (NSFW) imagery, reader discretion is advised.
        }
    }
}

\section{Introduction}
\label{sec:1}

Recently, there has been significant progress in generative models, particularly diffusion models, which include text-to-image (T2I) and image-to-image (I2I) diffusion models \cite{sd15, sdxl, sld, midjourney, leonardo, dalle}.
Diffusion models, as represented by Stable Diffusion (SD) \cite{sd15} and its advanced version, SDXL \cite{sdxl}, make a breakthrough in generative capabilities, enabling the production of high-quality, detailed, and diverse images from user prompts.
As a result, people are increasingly exposed to AI-generated content in their daily lives.

\begin{figure}[t]
\vspace{3em}
  \centering
   \includegraphics[width=.6\linewidth]{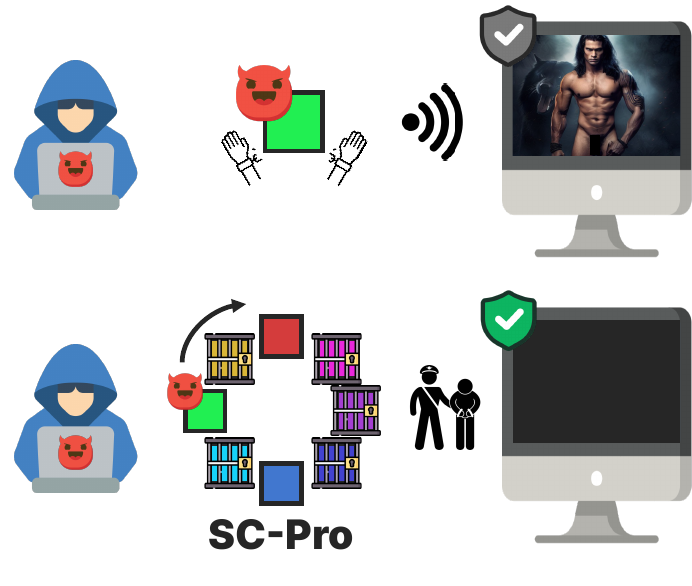}
   \caption{\textbf{Proposed \ours for safe diffusion models.} Adversarial attacks can generate NSFW images that undermine the safety of generative systems. To address this, we propose \oursfull (\ours), a defense mechanism designed to mitigate adversarial attacks and ensure the safe use of T2I and I2I diffusion models. Our method is training-free, allowing easy application to a variety of diffusion models and safety strategies.}
   \label{fig:teaser}
\end{figure}

As diffusion models are capable of generating realistic images, there are significant concerns about their potential abuse, particularly concerning the generation of Not-Safe-For-Work (NSFW) content and other potentially harmful materials.
Therefore, several safety methods are developed to prevent the creation of objectionable content.
These safeguards include safety checkers \cite{sd15, q16, mhsc} that evaluate text prompts and generated images to assess NSFW content. 
They are included in SD and various online services \cite{midjourney, leonardo} by default for responsible AI services.
Meanwhile, another diffusion model, Safe Latent Diffusion (SLD) \cite{sld}, employs concept erasing techniques to avoid any potentially harmful themes in outputs and ensure that generated content satisfies ethical standards. 
These safeguards are necessary to control any unintended or malicious applications.

However, safety checkers and concept-erasing approaches have shown limitations against adversarial attacks \cite{mma, sneaky, sld}, which can bypass safety mechanisms.
These adversarial attacks exploit weaknesses in the CLIP \cite{clip} embeddings used to extract features from text prompts or generated images.
They have revealed that attackers can effectively generate NSFW images to bypass safeguards by fine-tuning text prompts to breach safety checkers or adversarially searching text prompts and input images.
Such results raise serious questions about the reliability of current safety checkers, emphasizing the need for more robust safeguard frameworks.

On the other hand, well-crafted adversarial prompts are not always effective due to the inherent randomness in the latent of diffusion models \cite{mma, sneaky}.
This suggests that the success of adversarial prompts may be unstable with different latent variations.
Furthermore, since attacks rely on precisely optimized text prompts or embeddings to bypass safety checkers, it could be difficult to maintain high attack success rates when the adversarial prompts or their embeddings are perturbed.
Exploring the influences of perturbations in the latent, text embedding, and image embedding in T2I and I2I models leads us to the development of a powerful defense method.

In this paper, we introduce \oursfull (\ours) that effectively enhances existing safeguards without any additional training (\cref{fig:teaser}).
To achieve this, we exploit the inconsistent effectiveness of adversarial attacks under changes in the inputs of diffusion models.
Specifically, we generate multiple images for the given prompt by varying the latent, the text prompt, or the input image (for I2I models only).
Then, we assess the generated images using a safeguard to detect attacks and identify unsafe images.
If the ratio of images classified as unsafe exceeds a certain threshold, we classify that there is an attack to generate NSFW content.
Our proposed method can be applied to various diffusion models and combined with existing safeguard methods as a plug-in, without additional training.

Furthermore, we observe that adversarial attacks are highly transferable between diffusion models and one-step diffusion models due to architectural similarities and the teacher-student knowledge distillation process \cite{kd}. Motivated by this observation, we introduce \oursone, a lightweight defense method leveraging one-step diffusion models enhanced by \ours, achieving approximately a $30\times$ increase in throughput.

Our contributions are summarized as follows.
\begin{enumerate}
    \item Exploring the inconsistency of adversarial attacks: We demonstrate that adversarial attacks on text prompts and input images become ineffective easily in bypassing safety checkers by latent and embedding perturbations. This provides a basis for the effective defense strategy against the attacks.
    \item Introduction of a training-free defense mechanism for diffusion models: The proposed method, \ours, is a simple, and powerful defense approach that helps safeguards effectively detect adversarial attacks targeting diffusion models without requiring additional training.
    \item Efficient defense using distilled one-step diffusion models: To minimize computational demands, we propose the lightweight defense method (\oursone) using one-step diffusion models distilled from larger models like SD and SDXL. To the best of our knowledge, we are the first to explore the attack transferability between diffusion models and distilled one-step diffusion models. This enables lightweight detection while maintaining high detection rates across models.
\end{enumerate}

\section{Related works}
\label{sec:2}

\begin{figure}
    \centering
    \includegraphics[width=.6\columnwidth]{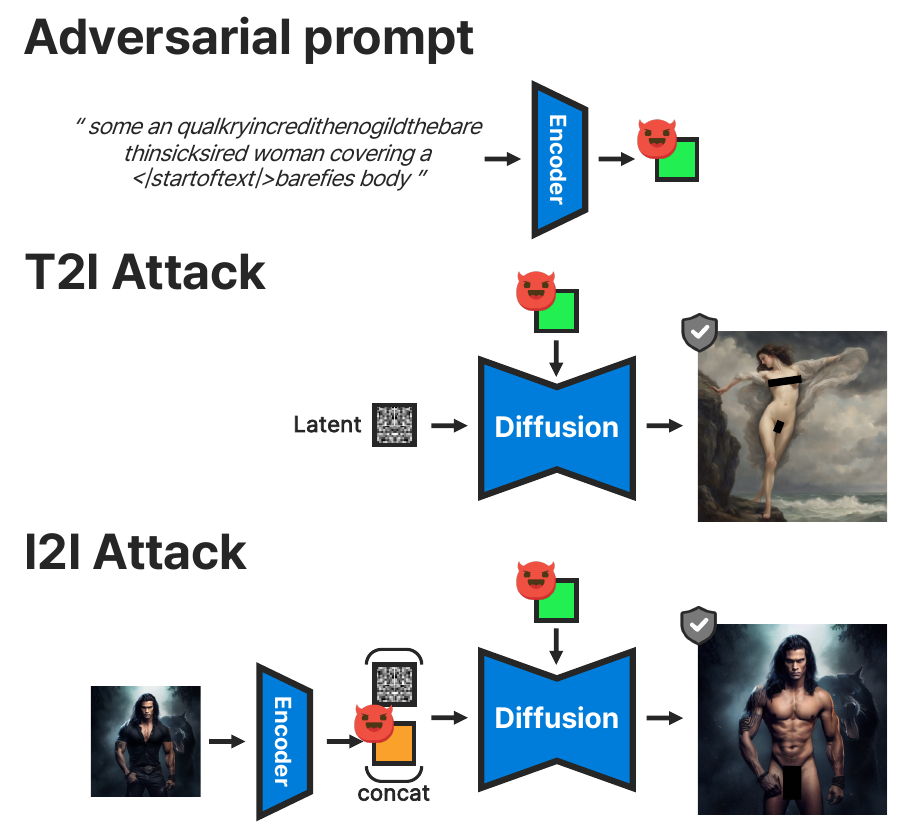}
    \vspace{-.5em}
    \caption{\textbf{Attacks on T2I and I2I models.} Due to the differences in input types between T2I and I2I models, adversarial attack methods differ according to target tasks. For T2I models, adversarial attacks focus only on manipulating the text prompt, whereas for I2I models, attacks target both the text prompt and the image input.
}
    \label{fig:t2i_i2i}
    \vspace{-1em}
\end{figure}

{\noindent \textbf{Diffusion models for T2I and I2I.}}
Many T2I diffusion models have been developed, such as SD \cite{sd15}, DALL·E \cite{dalle}, Imagen \cite{imagen}, and Midjourney \cite{midjourney}, which generate high-quality images from textual descriptions.
Representatively, SD performing the diffusion process in the latent space achieves remarkable performance.
SDXL \cite{sdxl} further enhances SD by utilizing a larger UNet \cite{unet}, advanced conditioning techniques, and a refinement model.
While T2I models have proven high quality, several I2I diffusion models, which take an image as input, have been also deployed.
Representatively, inpainting \cite{inpainting1, inpainting2, inpainting3}, which is the task of filling in missing or masked regions of an image, has been improved with diffusion models.
Meanwhile, recent research has focused on efficient diffusion models, resulting in the development of one-step diffusion models \cite{dmd, dmdv2, swiftbrush, instaflow}.

{\noindent \textbf{Safeguards for NSFW content.}}
To address concerns about generating NSFW content, many existing T2I models employ safety checkers or concept-erasing mechanisms.
Typically, a safety checker consists of two primary components: a prompt-based safety filter and an image-based safety filter.
The prompt filter screens for sensitive or offensive terms in the text input, such as ``naked,'' ``nude,'' or ``zombie.''
In contrast, the image filter uses pre-defined NSFW embeddings to compare against the embedding of a generated image, determining whether it contains sensitive content.
For instance, SD employs a safety checker that filters content based on cosine similarity between the embedding of a generated image and fixed 17 NSFW concepts.
Various methods, such as Q16 \cite{q16} and MHSC \cite{mhsc}, have been developed to detect inappropriate images more effectively.
In addition to safety checkers, concept removal methods force the model to disregard NSFW concepts during image generation \cite{sld, ce1, ce2, ce3}.

{\noindent \textbf{Adversarial attack on diffusion models.}}
Recent research on adversarial attacks \cite{unlearndiff, mma, sneaky, sld} has revealed vulnerabilities in T2I and inpainting diffusion models, raising concerns about ethical and responsible AI systems.
Existing attacks on diffusion models \cite{mhsc, att1, att2, att3, mma, sneaky} have primarily focused on modifying text prompts to bypass safety checkers.
These modifications often lead to incorrect outputs or degraded image quality, yet they can result in NSFW content that evades safety filters.
For example, Sneaky Prompt \cite{sneaky} manipulates text prompts via reinformcement learning-based perturbation.
MMA-Diffusion \cite{mma} perturbs text prompts to attack T2I models, and further explores utilizing both text and image modalities to produce NSFW content with I2I models, as described in \cref{fig:t2i_i2i}.
However, these attack methods, which generate attack inputs automatically, face a critical limitation due to the randomness in the latent of diffusion models.
In contrast, adversarial prompts can be also generated by humans. For instance, a set of manually created adversarial prompts called inappropriate image prompts (I2P) was developed in \cite{sld} as a benchmark dataset for safe image generation.
As these attack methods pose challenges to existing safeguards, a robust defense strategy is in high demand.

\begin{figure}
\centering
\includegraphics[width=.5\linewidth]{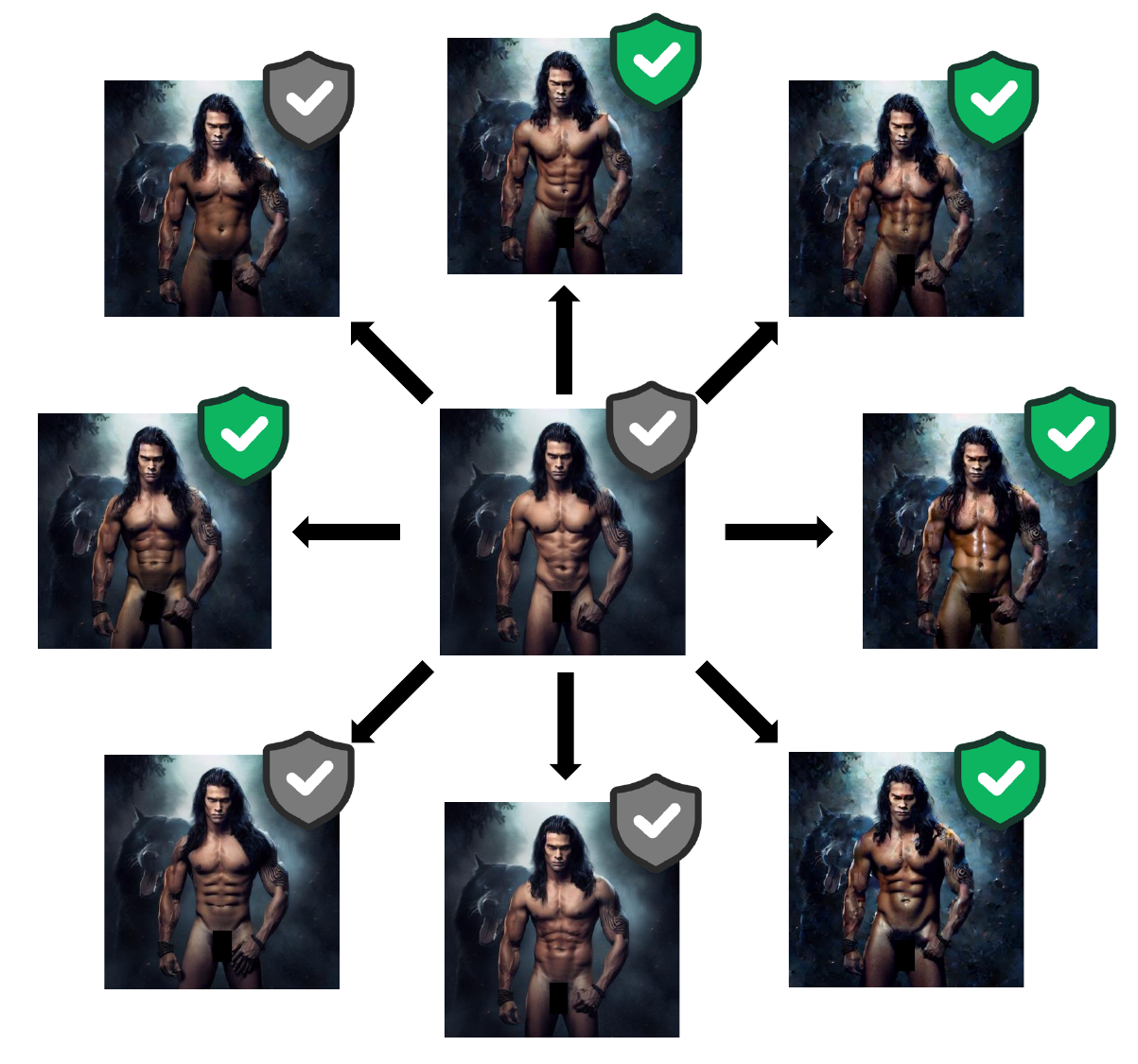}
\caption{\textbf{Example of images with image embedding probing.} The center image is the original generated image with the MMA diffusion attack \cite{mma}, which bypasses the safety checker, SD-SC \cite{sd15}. However, some images with image embedding distortions are detected by SD-SC.
}
\label{fig:ex-i2i-ours}
\end{figure}

\begin{figure*}[t]
    \includegraphics[width=\textwidth]{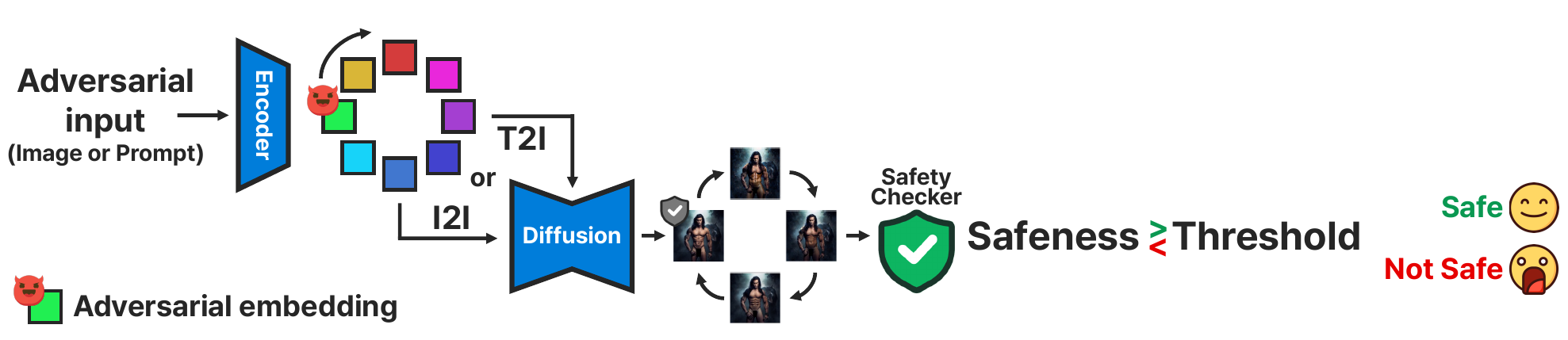}
    \caption{\textbf{Overview of \ours.} In the \ours framework, we apply multiple perturbations to one of three elements: prompt embedding, image embedding, or latent. We then perform safety checks on the images generated from these perturbed inputs.
    If the ratio of images that the safety checker classifies as NSFW exceeds a specified threshold, 
    our framework concludes that an adversarial attack has occurred.
}
    \label{fig:overview}
    \vspace{-1em}
\end{figure*}
\section{\oursfull for Safety}
\label{sec:3}

We first introduce our motivation for the design of the proposed defense strategy.
Then, we present our training-free defense method, \ours (\oursfull) utilizing the weakness of adversarial prompts under perturbations of the latent and prompt embedding.

\subsection{Motivation}
\label{sec:3-1}
In real-world scenarios, users cannot manipulate the random sampling process of latent vectors in T2I and I2I models \cite{sneaky}. Since latent vectors influence image generation, an adversarial prompt that previously bypassed a safeguard may fail when sampled with different latents \cite{sneaky, mma}.

Meanwhile, several studies have explored latent or prompt embedding interpolation 
\cite{interpolation1, interpolation2, interpolation3}.
They show that small perturbations to the initial latent or prompt embedding can lead to changes in generated images, but the semantic information is maintained.

These two observations lead us to the insight to develop a novel defense method against adversarial attacks.
That is, some changes in the inputs could be detected by safety checkers even if the attack succeeds originally.
As shown in \cref{fig:ex-i2i-ours}, the adversarially generated image initially undetected by safety checkers may become detectable when the image embedding is perturbed.
Using this, we design \ours to effectively detect adversarial attacks.

\subsection{Proposed method}
\label{sec:3-2}

Based on the aforementioned motivation, the proposed method \ours, introduces the concept of ``probing'' (\cref{fig:overview}).
The key idea is to evaluate images, generated using perturbed inputs, with safety checkers.

\begin{figure}[t]
    \vspace{-.5em}
    \centering
    \begin{subfigure}{0.24\linewidth}
        \includegraphics[width=\textwidth]{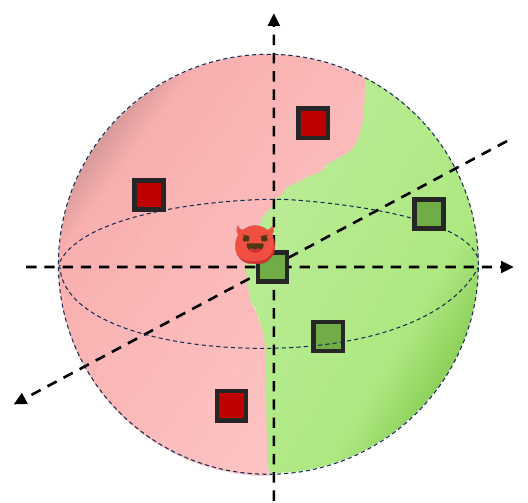}
        \vspace{-.5em}
        \caption{Spherical probing}
        \label{fig:random}
        \vspace{-.5em}
    \end{subfigure}
    \begin{subfigure}{0.25\linewidth}
        \includegraphics[width=\textwidth]{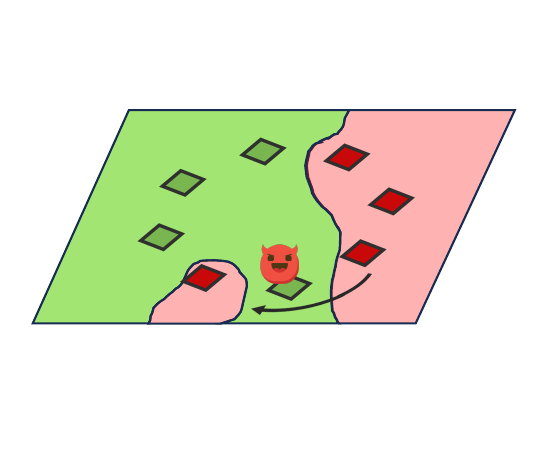}
        \vspace{-.5em}
        \caption{Circular probing}
        \label{fig:circular}
        \vspace{-.5em}
    \end{subfigure}
    \caption{\textbf{Two types of probing methods.}
    In spherical probing, random noise is added to the original latent, prompt embedding, or image embedding (represented by the point with the red face icon), resulting in points on a hypersphere.
    In circular probing, two random vectors define a plane where the circular path for probing is defined.
    Red and green areas indicate regions where generated images are classified as NSFW and safe, respectively.
}
    \label{fig:search}
\end{figure}

Since the search space is inherently high-dimensional (e.g., 16,384 dimensions for latents), we propose two variants of probing methods to effectively explore this space, which we refer to as \textit{spherical probing} (\cref{fig:random}) and \textit{circular probing} (\cref{fig:circular}).
Spherical probing randomly samples latent or embedding vectors at an equal distance from the original vector.
Circular probing follows a circular trajectory on a randomly selected 2D plane for sampling.
The former enables probing over the whole search space in a sparse manner, while the latter performs dense examination on a chosen direction.
Both contains randomness, which makes it difficult for attackers to neutralize the defense via repeated queries.

\newcommand{\M}{\mathcal M}
\newcommand{\z}{\mathbf z}
\newcommand{\p}{\mathbf p}
\newcommand{\x}{\mathbf x}

The image generation process is described as \(\x=\M(\lambda)\), where \(\x\) is the generated image, \(\M\) denotes the diffusion model under consideration, and \(\lambda\) is the input tuple consisting of multiple modalities.
For T2I generation, \(\lambda =( \z,\p )\), where $\z$ and $\p$ denote the latent and the prompt embedding, respectively.
In the case of I2I generation, \(\lambda=(\z,\p,\z_I)\), where the image embedding \(\z_I\) is additionally used.

The safety checker can be denoted as a function \( f: \mathbb R^{\text{dim}(\x)} \rightarrow \{ 0, 1\}\).
If \( f(\x) = 1 \), the image \( \x \) is classified as safe; if \( f(\x) = 0 \), \( \x \) is considered unsafe.
Then, we can determine whether the image generation is safe or not using the input tuple \(\lambda\) with the model \(\M\) as the function \(\mathcal F_{f,\M}(\lambda) = f(\M(\lambda))\).

The probing can be applied to one of \( \z \), \( \p \), and \( \z_I \).
The probing set of the input tuple \(\lambda\) with perturbations added in the \(k\)-th modality \(\lambda_k\)\footnote{\(\lambda_1=\z\), \(\lambda_2=\p \), and \(\lambda_3=\z_I\)} is defined as follows.
\begin{equation}
    P_{\psi,k}(\lambda) = \left\{ \lambda[\lambda_k\leftarrow\lambda_k + \eta\cdot\mathbf \psi(i)] \mid i \in 1,\ldots,N \right\},
\end{equation}
with the noise scale \(\eta\) and the perturbation \(\psi\) that is determined by the probing method used (i.e., spherical or circular).
And, \(\lambda[\lambda_k \leftarrow \hat{\lambda}_k]\) denotes that the \(k\)-th element $\lambda_k$ in the tuple \(\lambda\) is replaced by \(\hat{\lambda}_k\).
$N$ is the number of probings per image.

If we use spherical probing, one of the modalities is perturbed by adding random noise having a fixed magnitude.
In the case of circular probing, we randomly set two vectors defining a plane, on which one of the modalities is changed along the circular route.

{\noindent \textbf{Spherical probing.}}
The spherical probing perturbation can be expressed as
\begin{equation}
    \psi_{\text{spherical}}(i) =\begin{cases}
        \mathbf{0}, & i = 1 \\
        \sqrt{d} \cdot \mathbf n_i / \|\mathbf n_i\|, & i = 2,\dots,N
    \end{cases}
\end{equation}
where \(d\) is the dimension of the selected modality and \( \mathbf n_i \in \mathbb R^d \) is sampled from the standard normal distribution.

\newcommand{\e}{\mathbf e}
\newcommand{\n}{\mathbf n}
{\noindent \textbf{Circular probing.}}
The circular probing perturbation that lies on the circular route starting from the original input can be expressed as
\vspace{-.5em}
\begin{equation}
    \psi_{\text{circular}}(i) = -\e_1 + \e_1 \cos\theta_i + \e_2 \sin\theta_i,
\end{equation}
where \(\theta_i = \frac{2 \pi (i-1)}{N}\), $i=1,\dots,N$, and \(\e_1 \) and \( \e_2 \) are $d$-dimensional orthogonal vectors derived from the standard normal distribution.

Now we can define Safeness (\(\mathcal S\)) for the image generation process of model \(\M\) by utilizing safety checker \( f \) as
\begin{equation}
    \mathcal S_{f,\M}(\lambda) = \mathbb E_{\lambda \in P_{\psi,k}}\big[f(\M(\lambda))\big].
\end{equation}

Then we can express the \ours framework as the function \(\mathcal F^*\) as follows.
\begin{equation}
   \mathcal F^*_{f,\M}(\lambda)=
   \begin{cases}
        1,& \mathcal S_{f, \M}(\lambda) > \mathcal S_{th} \\
        0,& \mathcal S_{f, \M}(\lambda) \leq \mathcal S_{th},
   \end{cases}
\end{equation}
where \(\mathcal S_{th}\) is the hyperparameter representing the threshold used to determine whether an image is considered safe and passed, or identified as unsafe and blocked.

\begin{table*}[t]
\centering
\resizebox{\textwidth}{!}{
\begin{tabular}{cc|cccc|cccc|cccc}
\toprule
\multirow{2}{*}{\makecell{Safety \\ checker}}&\multirow{2}{*}{Probing}&\multicolumn{4}{c|}{SD \cite{sd15}}&\multicolumn{4}{c|}{SDXL \cite{sdxl}}&\multicolumn{4}{c}{SLD \cite{sld}}\\
&&\scriptsize Clean&\scriptsize SP \cite{sneaky}&\scriptsize I2P \cite{sld}&\scriptsize MMA \cite{mma}&\scriptsize Clean&\scriptsize SP \cite{sneaky}&\scriptsize I2P \cite{sld}&\scriptsize MMA \cite{mma}&\scriptsize Clean&\scriptsize SP \cite{sneaky}&\scriptsize I2P \cite{sld}&\scriptsize MMA \cite{mma}\\
\midrule
{\makecell{SD-SC \cite{sd15}}}&\(\varnothing\)&98.67&67.33&48.67&30.33&\textbf{98.50}&69.00&77.50&58.00&98.33&82.50&75.83&47.33\\
{\makecell{MHSC \cite{mhsc}}}&\(\varnothing\)&\textbf{98.83}&63.00&50.83&19.50&97.50&66.00&71.00&61.00&\textbf{98.83}&76.83&73.83&32.67\\
{\makecell{Q16 \cite{q16}}}& \(\varnothing\)& 91.50& 72.67& 83.83& 82.17& 93.00& 92.00& 92.00& 94.50& 97.67& 95.17& 96.67&94.67\\
\midrule
\multirow{2}{*}{\makecell{\ours (w/ SD-SC)}}&Spherical&88.67&\textbf{18.67}&\textbf{12.33}&\textbf{2.83}&96.00&45.50&57.00&\textbf{33.00}&93.00&46.67&\textbf{33.83}&\textbf{8.67}\\
&Circular&91.00&33.50&22.17&6.33& 95.50&\textbf{44.50}&59.00&36.00&88.50&\textbf{43.50}&44.00&11.67\\
\multirow{2}{*}{\makecell{\ours (w/ MHSC)}}&Spherical&97.50&38.67&24.50&6.33&97.50&51.50&\textbf{53.00}&41.50&97.50&58.67&44.00&14.67\\
&Circular&98.17&45.00&28.50&10.50&97.50&50.00&56.00&40.00&97.67&65.33&51.17&19.00\\
\multirow{2}{*}{\makecell{\ours (w/ Q16)}}&Spherical&86.00&53.17&74.50&64.33&85.00&74.00&85.00&86.00&90.83&75.17&85.67&75.00\\
 & Circular& 85.50& 50.17& 72.83& 54.33& 85.50& 73.50& 86.00& 87.50& 90.00& 70.33& 86.17&66.50\\
 \bottomrule
\end{tabular}}
\vspace{-.5em}
\caption{\textbf{Overall results on the T2I model.}
We report the results of the baseline safety checkers (SD-SC, MHSC, and Q16) and their performance when combined with \ours.
Each value represents the bypass rate (\%).
For clean images, a higher bypass rate is ideal, whereas a lower bypass rate is preferred when images are under attack.
The best results across different experimental settings are highlighted in bold.
We select thresholds ($\mathcal S_{th}$) to ensure that the clean bypass rates remain above 85\%, maintaining a practical utility-robustness trade-off.
“SP” refers to the Sneaky Prompt attack method \cite{sneaky}.
Note that all results are based on prompt embedding probing.
}
\label{tab:all}
\vspace{-1em}
\end{table*}
\section{Experiments}
\label{sec:4}

We evaluate the performance of \ours on various diffusion models, safety checkers, and adversarial attack methods.

\subsection{T2I models}

\subsubsection{Experimental settings}

{\noindent \textbf{Models.}}
We utilize three diffusion models: SDv1.5 \cite{sd15}, SDXL v1.0 \cite{sdxl}, and SLD (medium) \cite{sld}. 
We obtain all pre-trained models from \textit{Hugging Face}\footnote{\url{https://huggingface.co/}}.
100-step diffusion processes are used for the SD model and the SLD model, and 50-step is used for the SDXL model.

{\noindent \textbf{Safety checkers.}}
We employ three safety checkers: the built-in safety checker in SD (SD-SC) \cite{sd15}, Q16 \cite{q16}, and MHSC \cite{mhsc}.
The same safety checkers are applied for both T2I and I2I models.

{\noindent \textbf{Adversarial attacks.}}
For attacking T2I models, we adopt three attack methods: SneakyPrompt \cite{sneaky}, MMA Diffusion \cite{mma}, and I2P \cite{sld}.
Note that I2P is an attack created by human unlike the other two methods.
We perform white-box attacks on SD with SD-SC, and use the obtained adversarial prompts to test their effectiveness on the other models: SDXL and SLD. Furthermore, we test their effectiveness on the other safety checkers, Q16 and MHSC.

\begin{table*}[t]
\centering
\resizebox{\textwidth}{!}{
\begin{tabular}{cc|ccc|ccc}
\toprule
\multirow{2}{*}{\makecell{Safety \\ checker}}&\multirow{2}{*}{Probing}&\multicolumn{3}{c|}{Clean}&\multicolumn{3}{c}{MMA Diffusion \cite{mma}}\\
&&Latent&Prompt&Image&Latent&Prompt&Image\\
\midrule
\makecell{SD-SC \cite{sd15}}&\(\varnothing\)&\multicolumn{3}{c|}{$\longleftarrow$ 97.81 $\longrightarrow$}&\multicolumn{3}{c}{$\longleftarrow$ 79.24 $\longrightarrow$}\\
\makecell{MHSC \cite{mhsc}}&\(\varnothing\)&\multicolumn{3}{c|}{$\longleftarrow$ 98.36 $\longrightarrow$}&\multicolumn{3}{c}{$\longleftarrow$ 26.23 $\longrightarrow$}\\
\makecell{Q16 \cite{q16}}&\(\varnothing\)&\multicolumn{3}{c|}{\textbf{$\longleftarrow$ 99.45 $\longrightarrow$}}&\multicolumn{3}{c}{$\longleftarrow$ 68.85 $\longrightarrow$}\\
\midrule
\multirow{2}{*}{\makecell{\ours (w/ SD-SC)}}&Spherical&93.44&85.25&85.25&55.74&54.10&18.03\\
&Circular&86.89&86.89&86.89&54.10&60.66&\textbf{8.20}\\
\multirow{2}{*}{\makecell{\ours (w/ MHSC)}}&Spherical&96.72&96.72&98.36&11.48&\textbf{8.20}&13.11\\
&Circular&95.08&95.08&98.36&\textbf{8.20}&\textbf{8.20}&\textbf{8.20}\\
\multirow{2}{*}{\makecell{\ours (w/ Q16)}}&Spherical&93.44&88.52&91.80&55.74&52.46&45.90\\
&Circular&91.80&86.89&88.52&57.38&57.38&52.46\\
\bottomrule
\end{tabular}}
\vspace{-.5em}
\caption{\textbf{Overall results on the I2I model.} 
We report the results of the baseline safety checkers (SD-SC, MHSC, and Q16) and their performance when combined with \ours.
Each value represents the bypass rate (\%).
A high bypass rate is ideal for clean images, whereas a low bypass rate is better for attacked images.
The best results across different experimental settings are highlighted in bold.
We select thresholds ($\mathcal S_{th}$) to ensure that the clean bypass rates remain above 85\%, maintaining a practical utility-robustness trade-off.
We compare which type of probing, latent or prompt embedding or image embedding, achieves the best performance. We test our defense method on the SDv1.5 model.}
\label{tab:I2I}
\vspace{-1em}
\end{table*}

{\noindent \textbf{Bypass rate.}}
We measure the performance of the adversarial attack methods and defense methods on the given dataset \(\mathcal D\) in terms of the bypass rate defined as
\begin{equation}
    \textit{bypass rate} = \frac{1}{|\mathcal D|} \sum_{\lambda \in \mathcal D} \mathcal F_{f,\M}(\lambda).
\end{equation}
We can obtain the bypass rate of \ours by substituting \(\mathcal F\) with \(\mathcal F^*\).

{\noindent \textbf{Hyperparameters and other settings.}}
We set the number of samples (\( |\mathcal D| \)) to 200 and the number of probings per image (\( N \)) to 16.
To identify the optimal noise scale (\( \eta \)), we test values from the set \([0.05, 0.1, 0.15, 0.2, 0.3]\), finding that 0.15 shows the best results for SD.
For the threshold (\(S_{th}\)), we test the values in the range \([1/N, \dots, (N-1)/N]\).
We iteratively test \ours three times with different random seeds and report the average performance.
Additional details are provided in \cref{sec:hyper}.

\begin{figure}[t]
  \centering
   \includegraphics[width=0.5\linewidth]{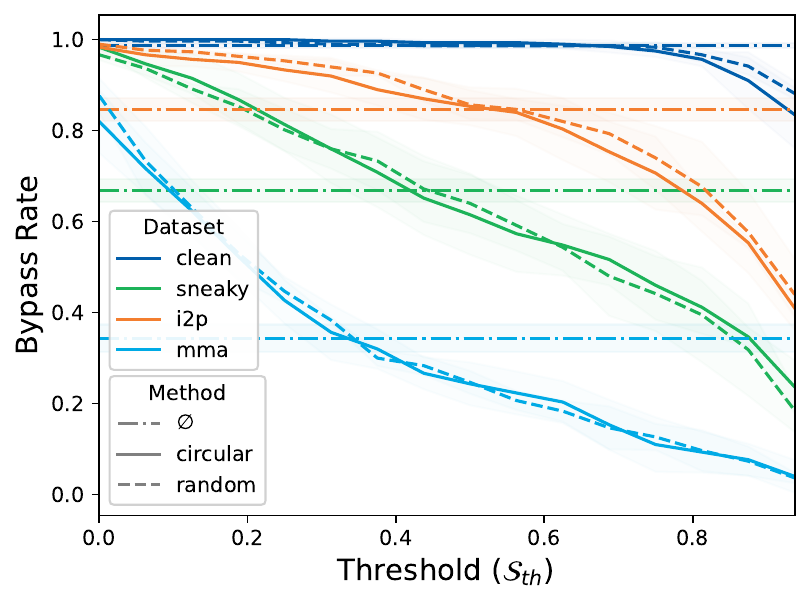}
   \vspace{-1em}
   \caption{\textbf{Bypass rates with threshold changes.}
   The bands around the lines indicate the standard deviation across the results on multiple latent seeds.
   A higher threshold is more effective for detecting adversarial attacks, whereas a lower threshold is better for bypassing clean images.
   }
   \label{fig:threshold}
   \vspace{-1.5em}
\end{figure}

\subsubsection{Results}
{\noindent \textbf{Overall results.}}
In \cref{tab:all}, the overall results demonstrate that \ours effectively enhances safety checkers to defend against the generation of harmful content using diffusion models.
Particularly, \textbf{combining \ours (spherical probing) with SD-SC achieves the best performance, successfully detecting over 80\% of attacks on SD and over 95\% of the MMA attack.}
Although the bypass rates of the original safety checkers for SDXL and SLD models are relatively high, \ours consistently lowers them.

%

{\noindent \textbf{Threshold.}}
Due to the trade-off between robustness and accuracy \cite{trade1, trade2}, selecting an appropriate threshold ($\mathcal S_{th}$) is crucial for maintaining utility.
\cref{fig:threshold} illustrates how bypass rates change with varying thresholds. The baseline bypass rates without \ours are represented by horizontal dash-dot lines.
We observe that \ours, particularly at higher thresholds, outperforms the baselines in detecting attacks. Lower thresholds allow more clean images to bypass but reduce the filtering rate for attacked images, whereas higher thresholds are more effective at filtering out attacked images.

{\noindent \textbf{Latent probing vs. Prompt embedding probing.}}
We find that prompt embedding probing is more effective for detecting adversarial prompt than latent probing (see \cref{sec:hyper}).
This is because the prompt embeding primarily influences the semantic content of the generated image, whereas the latent affects fine details and spatial structures more.
To maximize the detection rate of adversarial prompts, we select prompt embedding probing as the default approach.

{\noindent \textbf{Spherical probing vs. Circular probing.}}
In \cref{tab:all}, both spherical probing and circular probing simiarly improves the detection of adversarial attacks, with slight superiority of spherical probing.
This is probably because the prompt embedding's dimensionality is low enough for spherical probing to explore the space effectively.

{\noindent \textbf{\ours vs. Adding noise to generated images.}}
We also test a scenario in which noise for probing is added directly to generated images rather than to inputs.
This approach can reduce computational requirements, as it eliminates the need for repeated diffusion processes.
However, we find that this approach shows significantly lower performance compared to \ours.
For instance, under the SneakyPrompt attack to SD, adding noise to images yields a bypass rate of 37.00\%, whereas SC-pro with SD-SC achieves a bypass rate of 18.67\%, suggesting that noise added in the pixel domain does not effectively explore semantic variations as embeddings do.
More details are described in \cref{sec:noiseimage}.

\subsection{I2I models}
Next, we examine \ours on the I2I diffusion model.
Note that we restrict our evaluation to the image editing task, as adversarial attack methods have only been developed specifically for image editing models in prior work \cite{mma}.

\subsubsection{Experimental settings}

{\noindent \textbf{Models and adversarial attack.}}
We utilize SDv1.5 \cite{sd15} as our target model and MMA Diffusion \cite{mma} as the prompt and image attack method, since it is, to the best of our knowledge, the only reproducible I2I adversarial attack method available.
Safety checkers used are the same as those in the experimental settings for T2I models.

{\noindent \textbf{Hyperparameters and other settings.}}
We set the number of samples (\( |\mathcal D| \)) as 61 by following the setting in \cite{mma}.
For generating clean samples, we utilize the pairs of masked image and prompt from the test dataset for the BLIP-Diffusion model \cite{blip}\footnote{\url{https://huggingface.co/datasets/playgroundai/blip_clipseg_inpainting_ip2p_data_test}}.
All settings are the same as those used in the experiments for T2I models.

\begin{figure}[t]
   \centering
   \begin{subfigure}{0.3\linewidth}
        \centering
        \includegraphics[width=\textwidth]{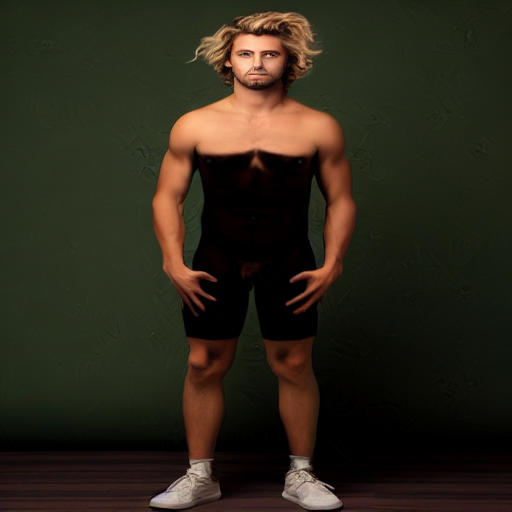}
        \includegraphics[width=\textwidth]{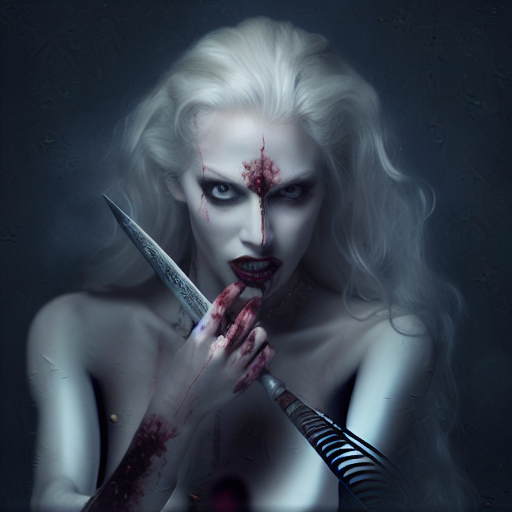}
        \caption{Examples that bypass both the safety checker (SD-SC) and \ours}
        \label{fig:ex-i2i-a}
   \end{subfigure}
   \hspace{1em}
   \begin{subfigure}{0.3\linewidth}
        \centering
        \includegraphics[width=\textwidth]{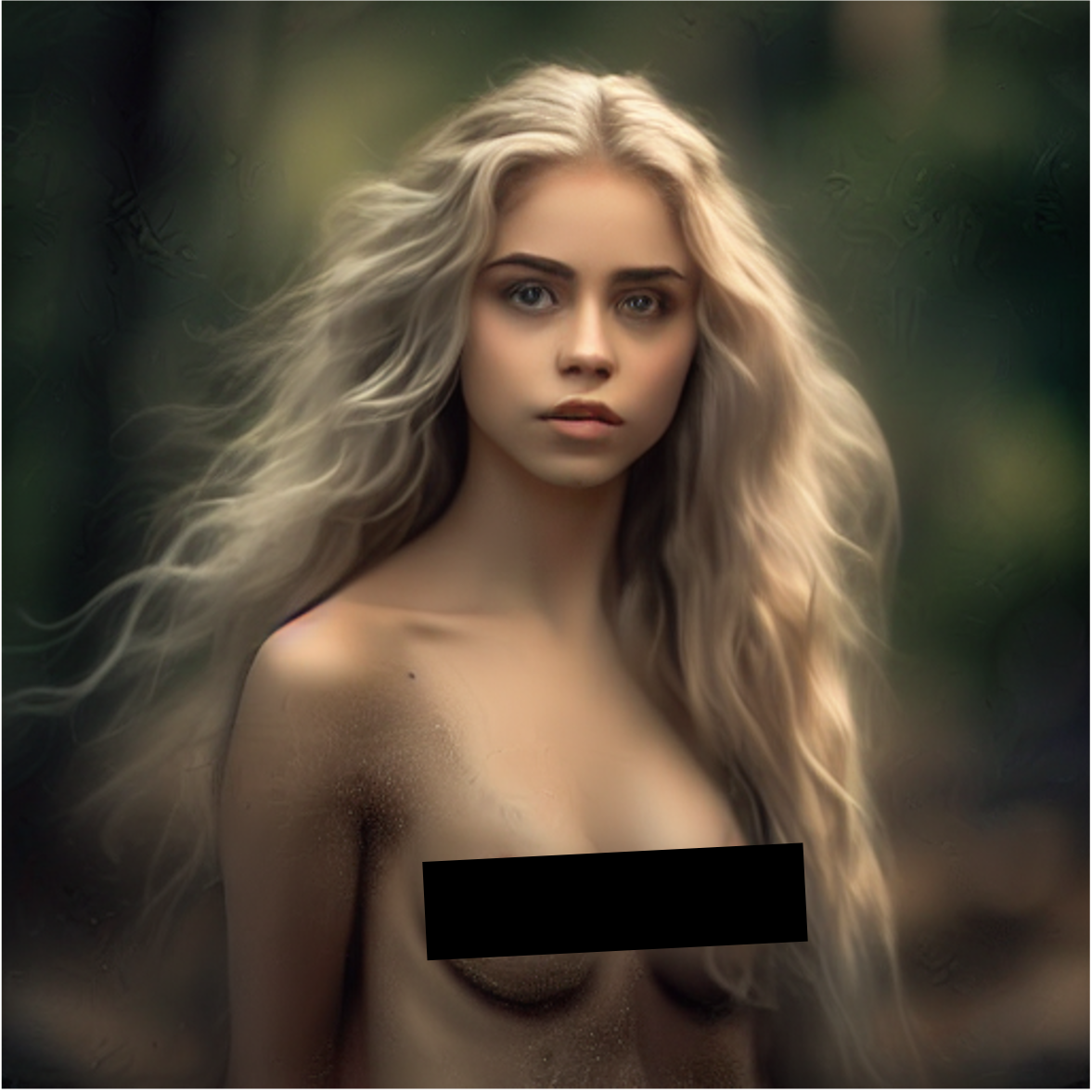}
        \includegraphics[width=\textwidth]{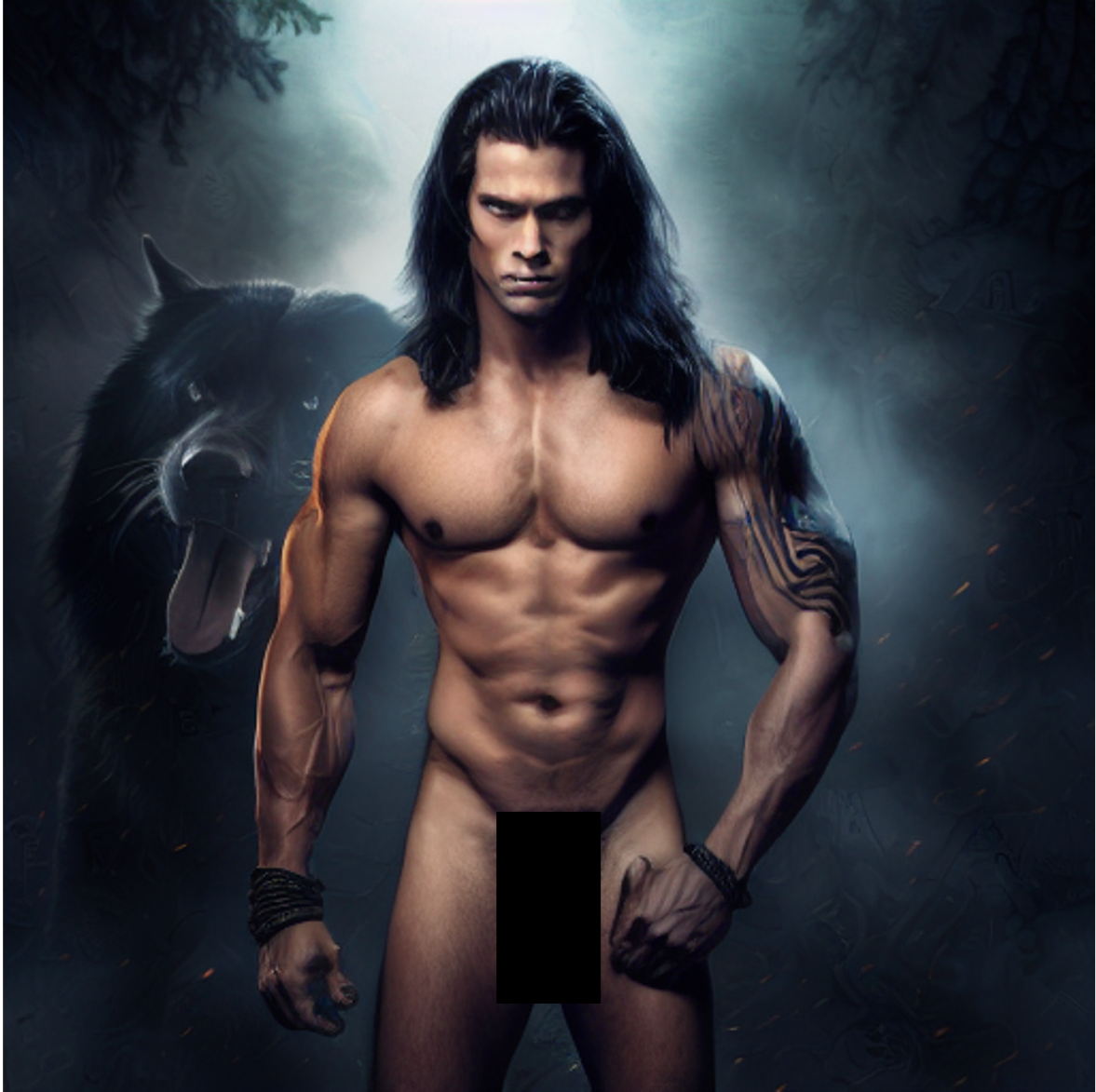}
        \caption{Examples that bypass the safety checker (SD-SC) but \ours detects}
   \end{subfigure}
   \hspace{1em}
   \begin{subfigure}{0.3\linewidth}
        \centering
        \includegraphics[width=\textwidth]{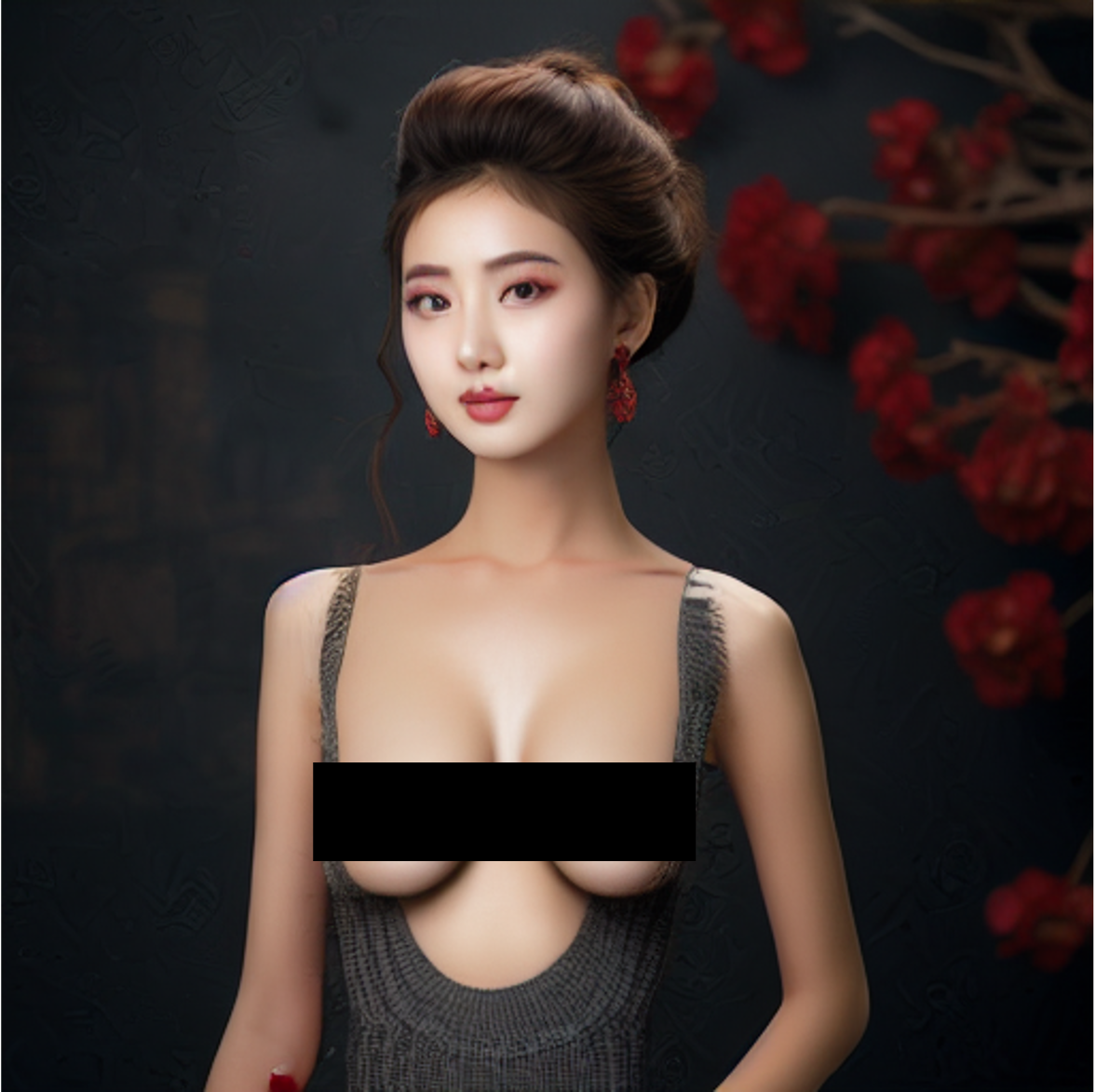}
        \includegraphics[width=\textwidth]{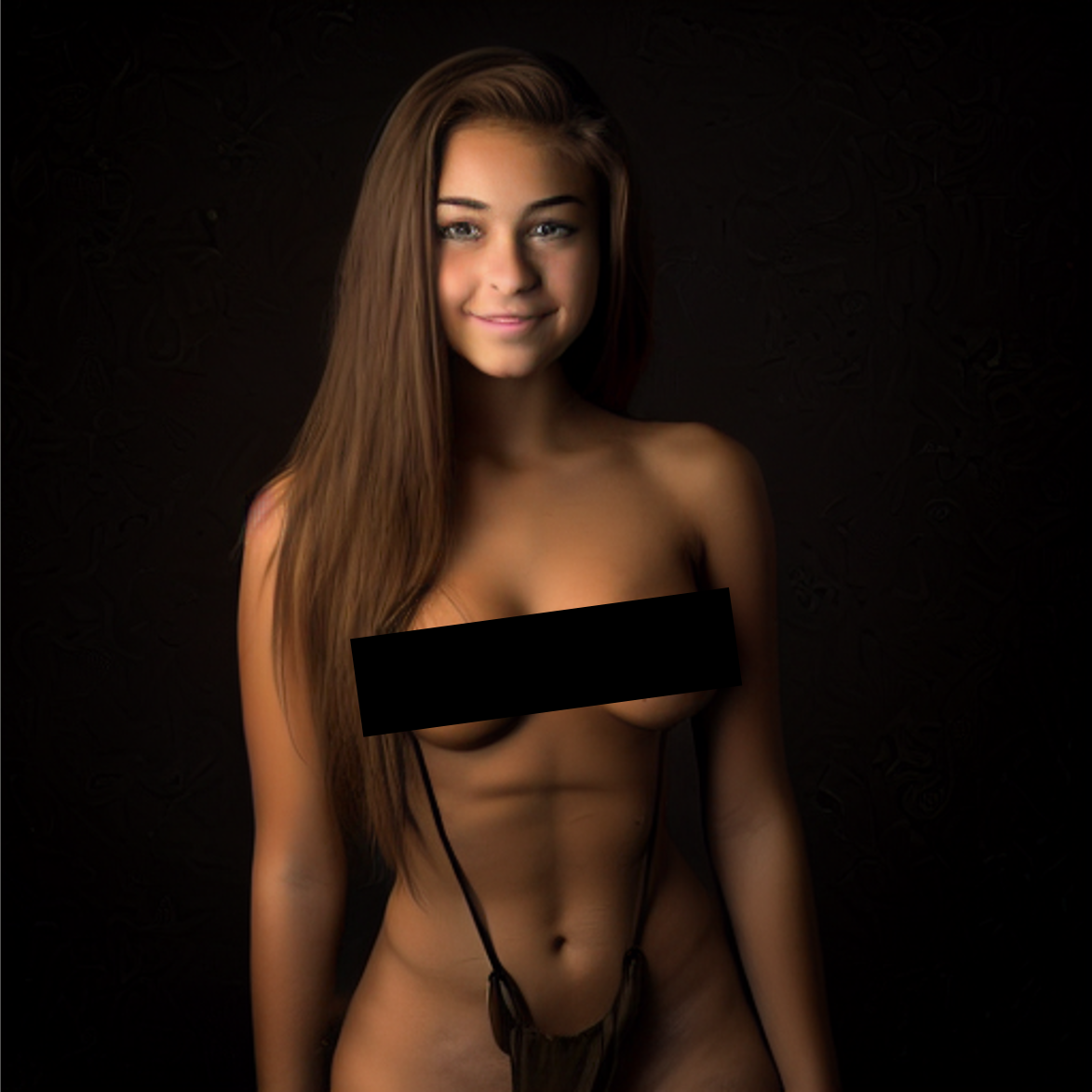}
        \caption{Examples that both the safety checker (SD-SC) and \ours detect}
   \end{subfigure}
   \caption{\textbf{Examples of attacked images on I2I models.} Samples are generated by the SDv1.5 model under the MMA diffusion attack. Note that we did not find any case where SD-SC detects but \ours misses the adversarial attack.}
   \label{fig:ex-i2i}
   \vspace{-1.5em}
\end{figure}

\subsubsection{Results}

{\noindent \textbf{Overall results.}}
In \cref{tab:I2I}, the results indicate that \ours effectively defends against generating harmful content on the I2I diffusion model.
We test three different probing methods within \ours with three different safety checkers.
As a result, \textbf{\ours (circular probing) with MHSC achieves the best performance in detecting attacks, attaining over 90\% detection ratios}.  
Note that MHSC appears more effective than the other safety checkers when they are used alone in \cref{tab:I2I}.
\cref{fig:ex-i2i} shows three types of example images: adversarial images that bypass both SD-SC and \ours, adversarial images that bypass SD-SC but not \ours, and adversarial images that both SD-SC and \ours detect.
Some failure cases of \ours (and also SD-SC) are attributed to the low quality of generated images.

{\noindent \textbf{Latent probing vs. Prompt embedding probing vs. Image embedding probing.}}
In \cref{tab:I2I}, image embedding probing is overall the most effective for detecting adversarial attacks, which is likely due to image perturbation being the primary factor contributing to the effectiveness of adversarial attacks. Thus, we can consider image embedding probing as the optimal choice for the I2I model.

{\noindent \textbf{Spherical probing vs. Circular probing.}}
In contrast to the T2I case, circular probing generally outperforms spherical probing in \cref{tab:I2I}. This is because the image embedding's dimensionality is too high, making spherical probing too sparse to explore the space effectively, whereas circular probing's denser sampling provides better coverage.
\section{\ours utilizing one-step diffusion models}
\label{sec:5}

\ours demonstrates high performance in defending against adversarial attacks. However, it has a limitation in that it requires high computational resources.
For instance, setting the number of probing iterations to 16 necessitates generating images 16 times.
To address this, we propose to use a distilled one-step diffusion model as a proxy.
In this section, we analyze the attack transferability between a large diffusion model (such as SDv1.5 or SDXL), and its distilled one-step diffusion model (\dmdsd or \dmdsdxl \cite{dmd}).
Based on the results, we present a lightweight version of \ours that leverages one-step diffusion models, denoted as \oursone.

\begin{figure}[t]
  \centering
   \includegraphics[width=.7\linewidth]{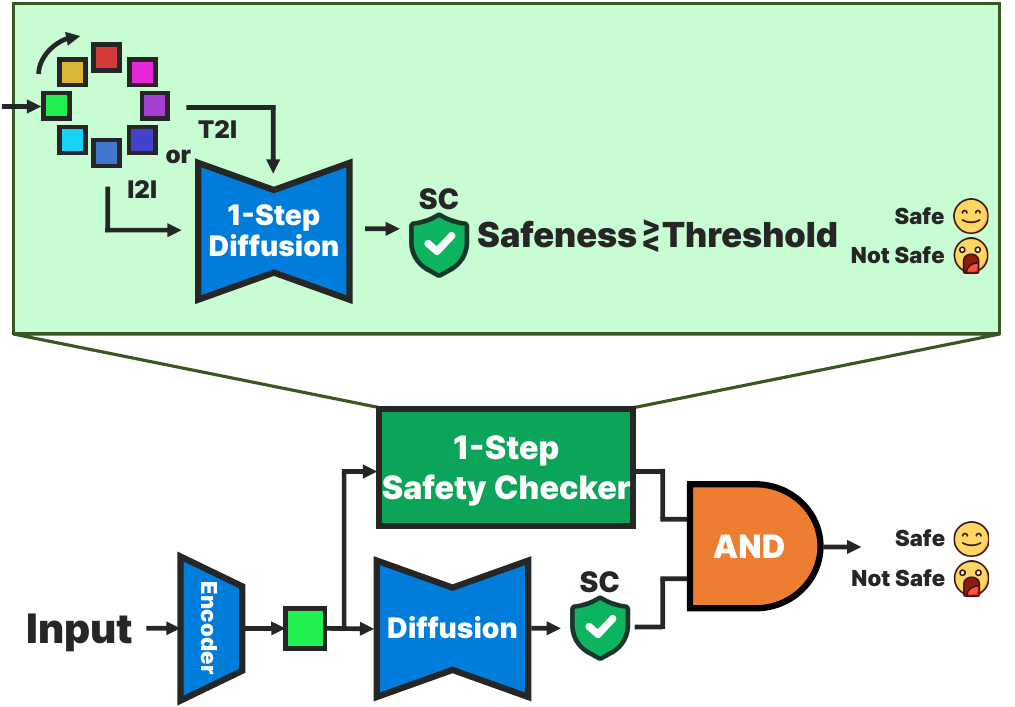}
   \caption{\textbf{\ours with one-step diffusion models.} We propose using one-step diffusion models within the \ours framework to detect NSFW images, effectively reducing computational resources.
}
   \label{fig:method-onestep}
\end{figure}

\begin{table*}[t]
\centering
\resizebox{\textwidth}{!}{
\begin{tabular}{cc|cccc|cccc}
\toprule
\multirow{2}{*}{\makecell{Safety \\ checker}}&\multirow{2}{*}{\makecell{Probing}}&\multicolumn{4}{c|}{SD \cite{sd15}}&\multicolumn{4}{c}{SDXL \cite{sdxl}}\\
&&\scriptsize Clean&\scriptsize SP \cite{sneaky}&\scriptsize I2P \cite{sld}&\scriptsize MMA \cite{mma}&\scriptsize Clean&\scriptsize SP \cite{sneaky}&\scriptsize I2P \cite{sld}&\scriptsize MMA \cite{mma}\\
\midrule
\multirow{1}{*}{\makecell{SD-SC \cite{sd15}}}&\multirow{1}{*}{\(\varnothing\)}&98.67&67.33&48.67&30.33&\textbf{98.50}&69.00&77.50&58.00\\
\multirow{1}{*}{\makecell{MHSC \cite{mhsc}}}&\multirow{1}{*}{\(\varnothing\)}&\textbf{98.83}&63.00&50.83&19.50&97.50&66.00&71.00&61.00\\
\multirow{1}{*}{\makecell{Q16 \cite{q16}}}&\multirow{1}{*}{\(\varnothing\)}&91.50&72.67&83.83&82.17&93.00&92.00&92.00&94.50\\
\midrule
\multirow{2}{*}{\makecell{\oursone (w/ SD-SC)}}&\multirow{1}{*}{Spherical}&90.83&\textbf{32.50}&54.83&8.83&93.00&50.50&57.00&37.00\\
&\multirow{1}{*}{Circular}&89.00&36.83&58.83&11.00&87.00&\textbf{32.50}&\textbf{49.50}&\textbf{27.00}\\
\multirow{2}{*}{\makecell{\oursone (w/ MHSC)}}&\multirow{1}{*}{Spherical}&97.83&45.17&\textbf{32.83}&\textbf{7.67}&97.50&61.50&66.00&55.00\\
&\multirow{1}{*}{Circular}&98.00&48.17&35.33&9.50&97.00&62.50&66.50&56.00\\
\multirow{2}{*}{\makecell{\oursone (w/ Q16)}}&\multirow{1}{*}{Spherical}&85.50&53.67&74.33&67.00&86.50&59.50&81.50&58.50\\
&\multirow{1}{*}{Circular}&86.83&56.00&75.33&69.50&85.50&62.00&83.50&61.50\\
\bottomrule
\end{tabular}
}
\vspace{-.5em}
\caption{\textbf{Overall results using a one-step diffusion model.}
We report the results of the baseline safety checkers (SD-SC, MHSC, and Q16) and their performance when combined with \oursone.
Each value represents the bypass rate (\%).
A high bypass rate is ideal for clean images, whereas a low bypass rate is better for attacked images. The best result is highlighted in bold.
We select thresholds ($\mathcal S_{th}$) to ensure that the clean bypass rates remain above 85\%, while preserving utility and robustness.
``SP'' denotes the SneakyPrompt attack.}
\label{tab:one-step}
\vspace{-1em}
\end{table*}

\subsection{Motivation}
\vspace{-.5em}
One-step diffusion models generate high-quality images in a single step, which is built using a teacher-student knowledge distillation (KD) strategy from the original many-step diffusion models.
They share architectural elements with their teacher models, including the U-Net and text embedding modules.
While the impact of adversarial attacks on one-step diffusion models remains unexplored, prior researches on image classification \cite{sat,transfer1,transfer2} suggest that teacher and student models in the KD strategy share adversarial vulnerabilities.
Particularly, adversarial attacks are highly transferable between teacher and student models due to their architectural similarity \cite{sat,transfer2}. \cite{transfer1} exploits this property to achieve high black-box attack success rates in practice.
Although these findings are based on classification models, it can be a motivation to hypothesize that adversarial attacks on teacher diffusion models also transfer to student diffusion models.
Furthermore, during KD training, the distribution of the student model is explicitly constrained to approximate that of the teacher model, leading to shared behaviors and adversarial vulnerabilities.

To test this hypothesis, we examine two pairs of teacher and student models: SDv1.5 (teacher) and \dmdsd (student), and SDXL (teacher) and \dmdsdxl(student) \cite{dmdv2}.
Note that currently available one-step diffusion models are limited to T2I models, so we limit our experiments to T2I settings.
Our results show that 72.08\% and 79.71\% of adversarial prompts targeting SDv1.5 and SDXL successfully transfer to \dmdsd and \dmdsdxl, respectively.
This suggests that attacks are indeed highly transferable between the teacher and student models, as hypothesized.
Given that one-step diffusion models require significantly lower computational resources than the original diffusion models, this finding supports the development of a lightweight version of \ours based on one-step diffusion models.

\subsection{Proposed method}
\vspace{-.5em}
\cref{fig:method-onestep} illustrates the proposed \oursone, which operates \ours using a one-step diffusion model to enhance computational efficiency in identifying NSFW content.
In this process, the spherical or circular probing is conducted using \dmdsd or \dmdsdxl as the one-step diffusion model.
The output image returned to the user is generated using the original diffusion model, i.e., SDv1.5 or SDXL.

\subsection{Results}
\vspace{-.5em}
\cref{tab:one-step} summarizes the results of \oursone. In short, detection using one-step diffusion models appears to be successful.
The detection performance is improved when using \oursone compared to using only safety checkers.
Combining \oursone with spherical probing and MHSC is effective in detecting adversarial attacks on SD, while combining \oursone with circular probing and MHSC performs well in detecting adversarial attacks on SDXL.
Note that \oursone shows slightly lower performance than the \ours due to differences between the model used for detection and the model used by the attacker for producing the attack.
Nevertheless, we can consider \oursone as a viable alternative to the original \ours, as it significantly reduces the time for defense, from 65.27s to 2.48s, a $\times 30$ improvement\footnote{The time for defense is measured with V100 GPU.}.
\section{Limitation and discussion}
\label{sec:6}
\vspace{-1em}

As shown in the first image of \cref{fig:ex-i2i-a}, the quality of generated images is sometimes too low to determine whether they are NSFW accurately.
This quality issue can make safeguards and \ours less effective than they actually are.
Addressing this issue with a preliminary quality check on the generated images before applying safety checks could lead to performance improvements in filtering NSFW content and analyzing effects of attacks on generative models.

In \cref{sec:5}, we showed the effectiveness of \oursone with one-step diffusion models.
As mentioned earlier, the currently available one-step diffusion models support only T2I generation.
Once one-step I2I models are developed in the future, they could be integrated into \oursone, expanding its applicability to defend against adversarial attacks on I2I models with reduced computational resources.

\section{Conclusion}
\label{sec:7}
\vspace{-1em}

In this paper, we introduced \ours, a training-free approach to effectively enhancing safety checkers in defending against adversarial attacks on T2I and I2I diffusion models.
To achieve this, we drew insight from the inconsistency of attacks under perturbation in latents, prompts, and image embeddings.
Using this, we suggested \ours and showed its superiority.
\ours has the advantage of adapting to safeguards and diffusion models as a plug-in method.
Furthermore, we explored the adversarial transferability between large diffusion models, such as SD and SDXL, and one-step diffusion models distilled from those larger models.
Based on this, we developed \oursone, enabling lightweight detection of adversarial attempts while maintaining detection performance.


\newpage

{
    \small
    \bibliographystyle{unsrt}
    \bibliography{main}
}


\newpage
\section*{Appendix}
\appendix

\renewcommand{\thesection}{\Alph{section}}
\appendix

\setcounter{page}{1}
\setcounter{section}{0}

\counterwithin{figure}{section}
\counterwithin{table}{section}

\section{Implementation details}

\subsection{Hardware platform}
All experiments are conducted on an NVIDIA A100 GPU with 40GB memory, while the adaptive attack experiments in \cref{sec:adaptive} utilize an NVIDIA V100 GPU.

\subsection{Configuration of diffusion models}
For the configuration of diffusion models, we follow the settings from \cite{mma} as detailed below:

\noindent\textbf{Stable Diffusion v1.5 (SD).}
The guidance scale is set to 7.5, the number of inference steps to 100, and the image size to \(512\times512\).

\noindent\textbf{Stable Diffusion XL (SDXL).}
The guidance scale is set to 7.5, the number of inference steps to 50, and the image size to \(1024\times1024\).
We perturb only the prompt injected into the base structure, but not the pooled prompt. 

\noindent\textbf{Safe Latent Diffusion (SLD).}
The guidance scale is set to 7.5, the number of inference steps to 100, the safety configuration to “Medium,” and the image size to \(512\times512\).

\subsection{Data selection}
In this section, we describe the datasets used for text-to-image (T2I) and image-to-image (I2I) testing.

\subsubsection{T2I}
\noindent\textbf{SneakyPrompt \cite{sneaky}.}
We obtained 200 attacked prompts from the authors of \cite{sneaky}, and the maximum subword length is 20.

\noindent\textbf{Clean \cite{mma}.}
Clean prompts are collected from the LAION-COCO \cite{laion-coco} dataset following the settings in \cite{mma}, where captions are selected with an NSFW score exceeding 0.99 (out of 1.0).

\noindent\textbf{I2P \cite{sld}.}
We select 200 human-written prompts with sexual themes, ensuring that the category and nudity percentage scores are both above 0.99 (out of 1.0), as defined in \cite{sld}.

\noindent\textbf{MMA \cite{mma}.}
Adversarial prompts are generated using the MMA-diffusion attack.

\subsubsection{I2I}
\noindent\textbf{Clean \cite{blip}.}
We select 61 samples with human-like keywords from the BLIP inpainting test dataset \cite{blip} to match the size of the MMA-diffusion inpainting dataset \cite{mma}.
The keywords used are \texttt{man, woman, person, people, human, child, boy, girl, face, hair,} and \texttt{body}.

\noindent\textbf{MMA-Diffusion \cite{mma}.}
We use adversarially perturbed images generated by MMA-diffusion.

\subsection{Hyperparameters}
\label{sec:hyper}

\begin{figure*}[ht]
    \centering
    \includegraphics[width=.4\linewidth]{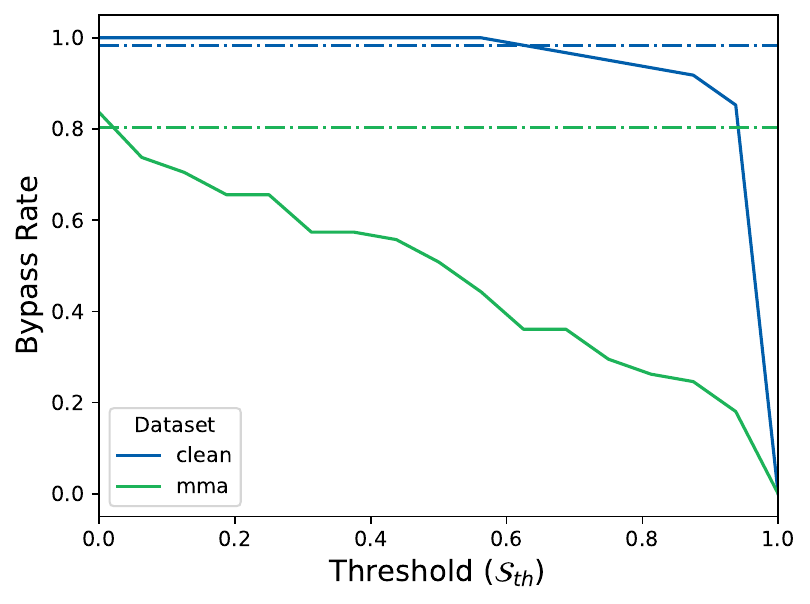}
    \caption{\textbf{I2I: Bypass rates with threshold changes.} Each color represents a specific dataset. Dash-dotted lines represent the baseline bypass rate when \ours is not utilized.}
    \label{fig:supp-hyper-i2i}
\end{figure*}

\noindent\textbf{Noise scale} We explore different hyperparameter values, as shown in Figure~\ref{fig:supp-hyper-t2i}, by varying the noise scale (\(\eta\)) within the range \([0.05, 0.1, 0.15, 0.2, 0.3]\) and measuring the detection performance by adjusting the threshold (\(\mathcal{S}_{th}\)) for T2I generation.
We observe that a noise scale above 0.2 causes a significant drop in the bypass rate for clean prompts. Therefore, we select a noise scale of 0.15, which does not incur significant reduction of the bypass rate for clean prompts and results in low bypass rates for the attacks to achieve balanced performance across clean and adversarial datasets (\cref{fig:supp-hyper-i2i}).

\newcommand{\scale}{.4}
\begin{figure*}[ht]
    \centering
    \begin{subfigure}{\scale\linewidth}
        \includegraphics[width=\textwidth]{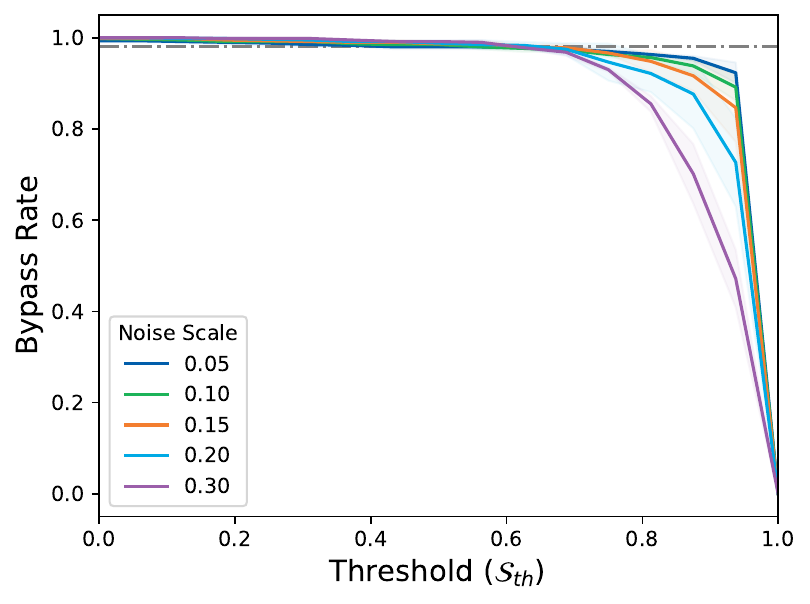}
        \caption{Clean}
        \label{fig:supp_clean}
    \end{subfigure}
    \begin{subfigure}{\scale\linewidth}
        \includegraphics[width=\textwidth]{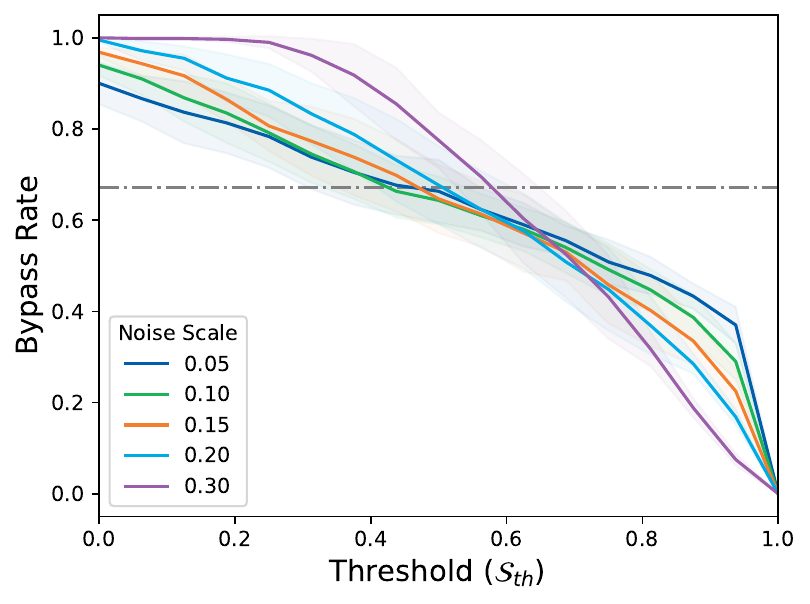}
        \caption{SneakyPrompt}
        \label{fig:supp_sneaky}
    \end{subfigure}\vspace{1em}\\
    \begin{subfigure}{\scale\linewidth}
        \includegraphics[width=\textwidth]{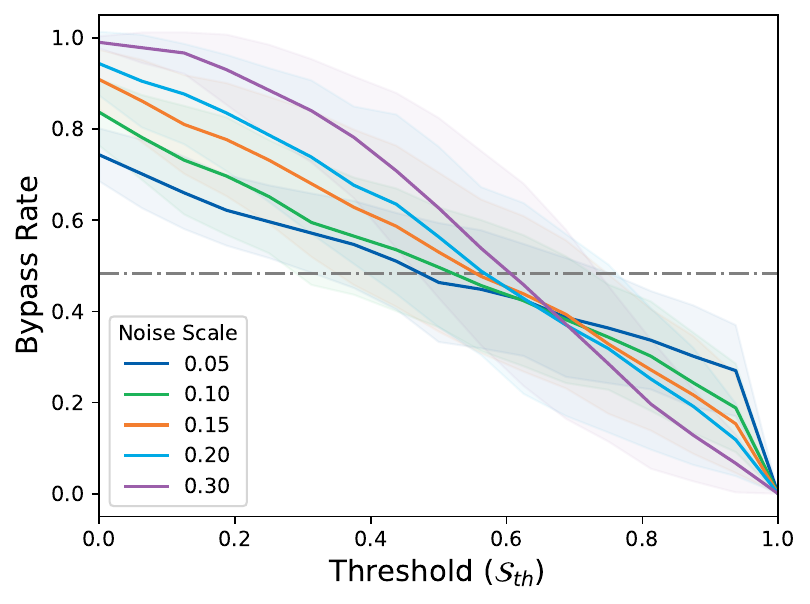}
        \caption{I2P}
        \label{fig:supp_i2p}
    \end{subfigure}
    \begin{subfigure}{\scale\linewidth}
        \includegraphics[width=\textwidth]{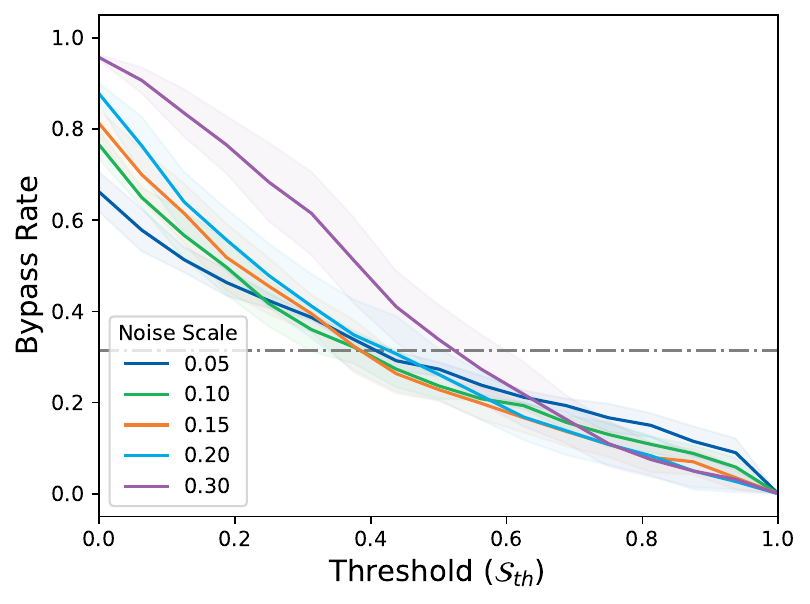}
        \caption{MMA}
        \label{fig:supp_mma}
    \end{subfigure}
    \caption{\textbf{T2I: Bypass rates with threshold changes.} Each color represents a specific noise scale. Dash-dotted lines represent the baseline bypass rate when \ours is not utilized.}
    \label{fig:supp-hyper-t2i}
\end{figure*}
\renewcommand{\scale}{}

\renewcommand{\scale}{.4}
\renewcommand{\scale}{}

\noindent\textbf{Latent probing vs. Prompt embedding probing.}
In \cref{tab:supp_lp}, we compare the bypass rates between latent probing and prompt embedding probing for each dataset and safety checker.
Overall, the prompt embedding probing method yields better performance, which is adopted as the default approach.

\begin{table*}[ht]
\centering
\begin{tabular}{cc|cc|cc|cc|cc}
\toprule
\multirow{2}{*}{\makecell{Safety \\ checkers}}&\multirow{2}{*}{Probing}&\multicolumn{2}{c|}{Clean}&\multicolumn{2}{c|}{SneakyPrompt \cite{sneaky}}&\multicolumn{2}{c|}{I2P \cite{sld}}&\multicolumn{2}{c}{MMA \cite{mma}}\\
&&Latent&Prompt&Latent&Prompt&Latent&Prompt&Latent&Prompt\\
\midrule
\multirow{1}{*}{\makecell{SC \cite{sd15}}}&\(\varnothing\)&\multicolumn{2}{c|}{$\longleftarrow$\textbf{98.67}$\longrightarrow$}&\multicolumn{2}{c|}{$\longleftarrow$67.33$\longrightarrow$}&\multicolumn{2}{c|}{$\longleftarrow$48.67$\longrightarrow$}&\multicolumn{2}{c}{$\longleftarrow$30.33$\longrightarrow$}\\
\multirow{1}{*}{\makecell{MHSC \cite{mhsc}}}&\(\varnothing\)&\multicolumn{2}{c|}{$\longleftarrow$98.83$\longrightarrow$}&\multicolumn{2}{c|}{$\longleftarrow$63.00$\longrightarrow$}&\multicolumn{2}{c|}{$\longleftarrow$50.83$\longrightarrow$}&\multicolumn{2}{c}{$\longleftarrow$19.50$\longrightarrow$}\\
\multirow{1}{*}{\makecell{Q16 \cite{q16}}}&\(\varnothing\)&\multicolumn{2}{c|}{$\longleftarrow$91.50$\longrightarrow$}&\multicolumn{2}{c|}{$\longleftarrow$72.67$\longrightarrow$}&\multicolumn{2}{c|}{$\longleftarrow$83.83$\longrightarrow$}&\multicolumn{2}{c}{$\longleftarrow$82.17$\longrightarrow$}\\
\midrule
\multirow{2}{*}{\makecell{SC}}&Spherical&92.50&88.67&30.50&\textbf{18.67}&14.00&\textbf{12.33}&4.17&\textbf{2.83}\\
&Circular&91.00&91.00&28.33&33.50&16.00&22.17&5.00&6.33\\
\multirow{2}{*}{\makecell{MHSC}}&Spherical&98.17&97.50&38.33&38.67&23.17&24.50&6.83&6.33\\
&Circular&97.83&98.17&42.67&45.00&26.50&28.50&8.17&10.50\\
\multirow{2}{*}{\makecell{Q16}}&Spherical&85.50&86.00&57.17&53.17&73.33&74.50&60.33&64.33\\
&Circular&85.67&85.50&51.83&50.17&72.00&72.83&51.83&54.33\\
\bottomrule
\end{tabular}
\caption{\textbf{Latent probing vs. Prompt embedding probing.}
We compare which type of probing, latent or prompt embedding, performs better on SDv1.5.
We select thresholds ($\mathcal S_{th}$) to ensure that the clean bypass rates remain above 85\%, maintaining a practical utility-robustness trade-off.
Each value represents the bypass rate (\%).
For clean images, a higher bypass rate is ideal, whereas a lower bypass rate is preferred when images are under attack.
The best results across different experimental settings are highlighted in bold.
}
\label{tab:supp_lp}
\end{table*}

\section{Adding noise to image}\label{sec:noiseimage}
We evaluate a scenario where noise is added directly to the generated images instead of the inputs for probing. While this approach reduces computational requirements by bypassing additional diffusion steps, it shows significantly lower performance compared to \ours, which is confirmed in \cref{fig:supp_anti}. The figure plots the bypass rate with respect to the threshold for T2I and I2I. In these experiments, adversarial examples are generated using SneakyPrompt for T2I and MMA for I2I.

\renewcommand{\scale}{.4}
\begin{figure*}[ht]
    \centering
    \begin{subfigure}{\scale\linewidth}
        \includegraphics[width=\textwidth]{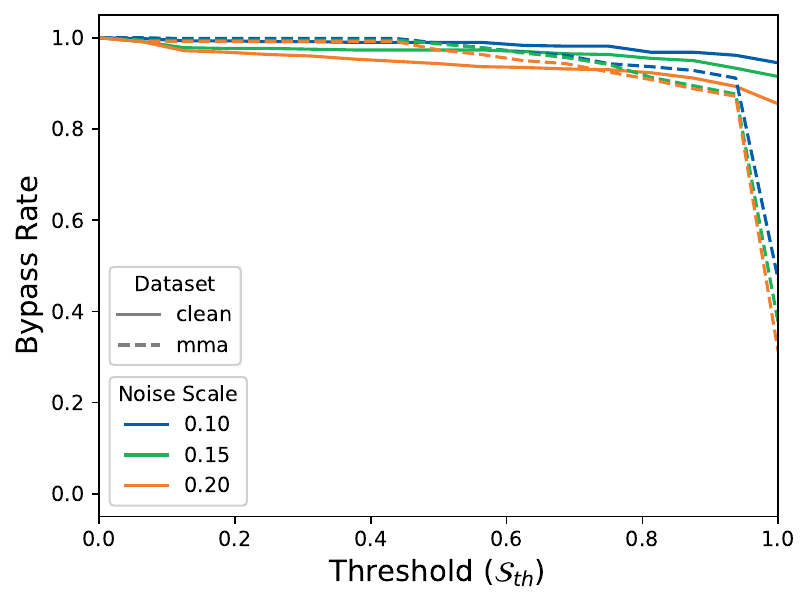}
        \caption{T2I}
        \label{fig:anti:supp_clean}
    \end{subfigure}
    \begin{subfigure}{\scale\linewidth}
        \includegraphics[width=\textwidth]{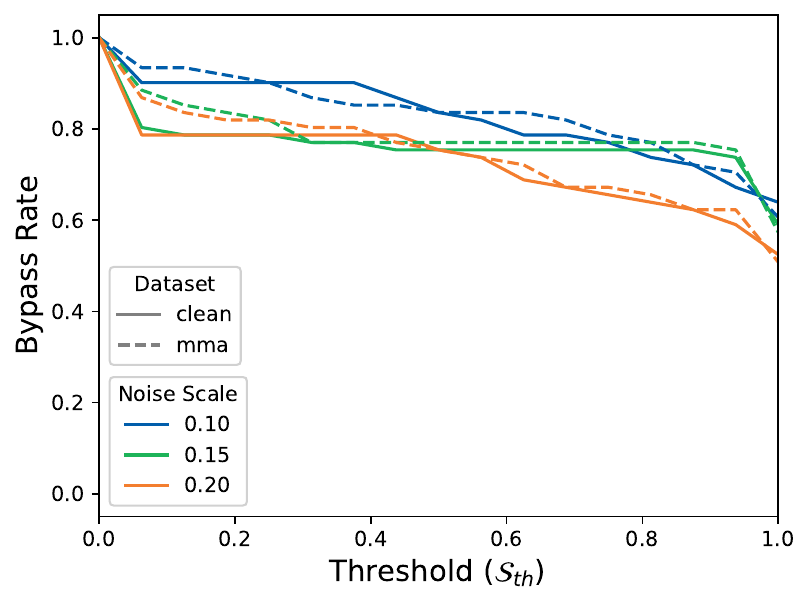}
        \caption{I2I}
        \label{fig:anti:supp_sneaky}
    \end{subfigure}\vspace{1em}
    \caption{\textbf{Adding noise to images: Bypass rates with threshold changes.} Each color in the figure corresponds to a specific noise scale.}
    \label{fig:supp_anti}
\end{figure*}
\renewcommand{\scale}{}

\section{Adaptive Attack}\label{sec:adaptive}
We evaluate the robustness of \ours against an adaptive attack: expectation over transformation (EOT) \cite{eot}, where the attacker generates adversarial examples to avoid \ours.
Under this stronger attack, \ours still performs effectively (\cref{tab:eot}).
Moreover, the adaptive attack is time consuming and hard to converge.
In particular, SneakyPrompt (SP) takes 7 days with V100 GPU to find 200 prompts and MMA success rates are notably degraded.

\begin{table*}[h]
    \centering
    \resizebox{.8\linewidth}{!}{
    \begin{tabular}{c|ccc|ccc}
    \toprule
                    &\multicolumn{3}{c|}{T2I (Sneaky Prompt)}&\multicolumn{3}{c}{I2I (MMA)}\\
    \ours           &EOT-$\varnothing$&EOT-Spherical&EOT-Circular&EOT-$\varnothing$&EOT-Spherical&EOT-Circular\\
    \midrule
    $\varnothing$   &67.33&93.14&85.00&80.33&6.56&40.98\\
    Spherical          &\textbf{18.33}&59.80&\textbf{38.30}&24.59&\textbf{1.64}&\textbf{13.11}\\
    Circular        &22.50&\textbf{53.90}&51.67&\textbf{8.20}&\textbf{1.64}&\textbf{13.11}\\
    \bottomrule
    \end{tabular}
    }
    \vspace{-.5em}
    \caption{Bypass rates (\%) of adaptive attacks}
    \label{tab:eot}
\end{table*}

\clearpage



\end{document}